\documentclass{article}

\PassOptionsToPackage{numbers}{natbib}

\usepackage[final]{neurips_2019}

\usepackage[utf8]{inputenc} 
\usepackage[T1]{fontenc}    
\usepackage{hyperref}       
\usepackage{url}            
\usepackage{booktabs}       
\usepackage{amsfonts}       
\usepackage{nicefrac}       
\usepackage{microtype}      

\usepackage{enumitem}

\usepackage{graphicx}
\usepackage{subcaption}
\usepackage{appendix}

\usepackage{mathtools}
\usepackage{amsmath}
\usepackage{siunitx} 

\usepackage{todonotes}

\title{Spatial and Colour Opponency in Anatomically Constrained Deep Networks}

\author{%
  Ethan Harris\thanks{Authors contributed equally}\hspace{0.4em}\thanks{Vision, Learning and Control Group, Electronics and Computer Science, University of Southampton, \{ewah1g13, adm1g15, jsh2\}@ecs.soton.ac.uk}\\
  \And
  Daniela Mihai\footnotemark[1]\hspace{0.4em}\footnotemark[2]\\
  \And
  Jonathon Hare\footnotemark[1]\hspace{0.4em}\footnotemark[2]\\
}

\begin{document}

\maketitle

\begin{abstract}
Colour vision has long fascinated scientists, who have sought to understand both the physiology of the mechanics of colour vision and the psychophysics of colour perception. We consider representations of colour in anatomically constrained convolutional deep neural networks. Following ideas from neuroscience, we classify cells in early layers into groups relating to their spectral and spatial functionality. We show the emergence of single and double opponent cells in our networks and characterise how the distribution of these cells changes under the constraint of a retinal bottleneck. Our experiments not only open up a new understanding of how deep networks process spatial and colour information, but also provide new tools to help understand the black box of deep learning. The code for all experiments is avaialable at \url{https://github.com/ecs-vlc/opponency}.
\end{abstract}

\section{Introduction}

Opponent colour theory, which considers how combinations of chromatic stimuli are encoded in the visual system, proposed by \citet{hering1964outlines}, initiated nearly a century earlier by \citet{goethe1840theory}, was observed and formulated at a cellular level only in the 1950s by \citet{de1958response} and others \cite{wagner1960opponent,wiesel1966spatial,naka1966s,daw1967goldfish}. Combined, the theories of colour opponency, trichromacy \citep{young1802ii,helmholtz1852lxxxi,maxwell1860iv} and feature extraction in the visual cortex \citep{kuffler1953discharge,hubel1962receptive,hubel2004brain,troy2002receptive} constitute a deep understanding of early visual processing in nature. Furthermore, the notional elegance of these theories has served to motivate much of the progress made in computer vision, most notably including the development of Convolutional Neural Networks (CNNs) \citep{le1990handwritten,bottou1994comparison,lecun1995convolutional} that are now so focal in our collective interests. Despite the sheer volume of these experimental discoveries, they still represent only a sparse view of the broad spectra of the natural world. This limits our ability to consider precisely which physiological differences lead to the subtle variations in visual processing between species. For this reason, deep learning offers a unique platform through which one can study the emergence of distinct visual phenomena, across the full gamut of constraints and conditions of interest.

\citet{lindsey2019unified} use a multi-layer CNN to explore how the emergence of centre-surround and oriented edge receptive fields changes under biologically motivated constraints. Primarily, the authors find that the introduction of a bottleneck on the number of neurons in the second layer (the `retina' output) of a CNN (trained to classify greyscale images) induces centre-surround receptive fields in the early layers and oriented edges in the later layers. Furthermore, the authors demonstrate that as this bottleneck is decreased, the complexity of early filters increases and they tend towards orientation selectivity. The nature of colour in CNNs has also been explored \citep{engilberge2017color,gomez2018convolutional}. Specifically, \citet{engilberge2017color} find that spectral sensitivity is highest in the early layers and traded for class sensitivity in deeper layers. \citet{gomez2018convolutional} demonstrate that CNNs are susceptible to the same visual illusions as those that fool human observers. This lends weight to the notion that the specifics of colour processing result from our experience of visual stimuli in the natural world \citep{purves2011we}.

In this paper we find evidence for spectral and spatial opponency in a deep CNN with a retinal bottleneck (following \citet{lindsey2019unified}) and characterise the distribution of these cells as a function of bottleneck width. In doing so, we introduce a series of experimental tools, inspired by experiments performed in neurophysiology, that can be used to shed light on the functional nature of units within deep CNNs. Furthermore, we show similarities between the specific excitatory and inhibitory responses learned by our network and those observable in nature. Across all experiments our key finding is that structure (the separation of functional properties into different layers) emerges naturally in models that feature a bottleneck. Code, implemented using PyTorch~\citep{paszke2017automatic} and Torchbearer~\citep{torchbearer2018} is avaialable at \url{https://github.com/ecs-vlc/opponency}.

\section{Spatial and Colour Opponency in the Brain}
Experiments using micro-electrode recording have been used to explore how single cells respond to different stimuli. Consequently, a number of different observations and subsequent classifications of cells regarding behavioural characteristics have been made.
The first key observation is the existence of two types of cell that respond to colour; spectrally opponent and spectrally non-opponent. Cells with opponent spectral sensitivity~\citep{DeValois:66} are excited by particular colours,\footnote{Technically, the original experiments by~\citet{DeValois:66} used energy-normalised single-wavelength stimuli rather than a more general notion of colour created from a mixture of wavelengths.} and inhibited by others. For a cell to be inhibited its response must fall below its response to an empty stimulus (the `background rate'). For excitation to occur, the response must be at some point above the background rate. Additionally, \citet{DeValois:66} discovered that broadly speaking
the cells could be grouped into those that were excited by red and inhibited by green (and vice-versa), and cells that were excited by blue and inhibited by yellow (and vice-versa). Cells that are spectrally non-opponent are not sensitive to specific wavelengths (or colours of equal intensity) and respond to all wavelengths in the same way
. A second key observation is the existence of cells with spatial receptive fields that are opponent to each other; that is, in some spatial area, they are excited above the background rate by certain stimuli, and in other areas they are inhibited by certain stimuli \citep{de1958response}.
Cells responsive to colour can be further grouped into `single opponent' and `double opponent' cells.
These cells respond strongly to colour patterns but are only weakly responsive to full-field colour stimuli (e.g. solid colour across the receptive field, slow gradients or low frequency changes in colour)~\cite{SHAPLEY2011701}.

\section{Experiments}\label{sec:experiments}
%

In this section we detail our experimental procedures and results, characterising the emergence of spectrally, spatially and double opponent cells in deep CNNs. We focus here on the whole population of cells, for a depiction of the characterisation of a single cell see Appendix~\ref{app:single}. To preserve similarity with \citet{lindsey2019unified}, we adopt the same deep convolutional model of the visual system.
This model consists of two parts: a model of the retina, built from a pair of convolutional network layers with ReLU nonlinearities, and termed `retina-net'; and, a ventral stream network (VVS-net) built from a stack of convolutional layers (again with ReLU) followed by a two layer MLP (with 1024 ReLU neurons in the hidden layer, and a 10-way softmax on the output layer). All convolutions are 9x9 with same padding, and each has 32 channels, with the exception of the second retinal layer whose number of channels is the retinal bottleneck. The number of convolutional layers in the ventral stream is also a parameter of the model. Our visual system model is trained with the same range of parameters (varying retinal bottlenecks and ventral system depths), with the same optimisation hyperparameters as \citet{lindsey2019unified}, differing only in that it takes 3-channel colour inputs.
As with \citeauthor{lindsey2019unified}'s work, the networks are trained to perform classification on the CIFAR-10 dataset \citep{Krizhevsky09learningmultiple}.
Error bars throughout our experiments denote the standard deviation in result across all 10 models trained for each set of hyper-parameters. For further details see Appendix \ref{app:details}.
\paragraph{Spectral opponency}
%
To classify cells according to their spectral opponency, we can simulate the experimental procedure of \citet{DeValois:66}. Specifically, we first present the network with uniform coloured images and measure the response of the target cell. By sampling colour patches according to hue (see Appendix~\ref{app:colorpatches}) we can show the network a range of stimuli and construct a response curve. We then classify each cell as either `spectrally opponent' or `spectrally non-opponent' by considering this curve relative to a background rate, defined as the response of the cell to a zero image. A spectrally non-opponent cell is one for which all responses are either above or below the baseline. A spectrally opponent cell is one for which the response is above the baseline for some colours and below the baseline for others. We further define an additional class, spectrally unresponsive, for cells which respond the same regardless of the hue of the input.

%
\begin{figure}
    \centering
    \begin{subfigure}{0.32\linewidth}
        \includegraphics[width=\linewidth]{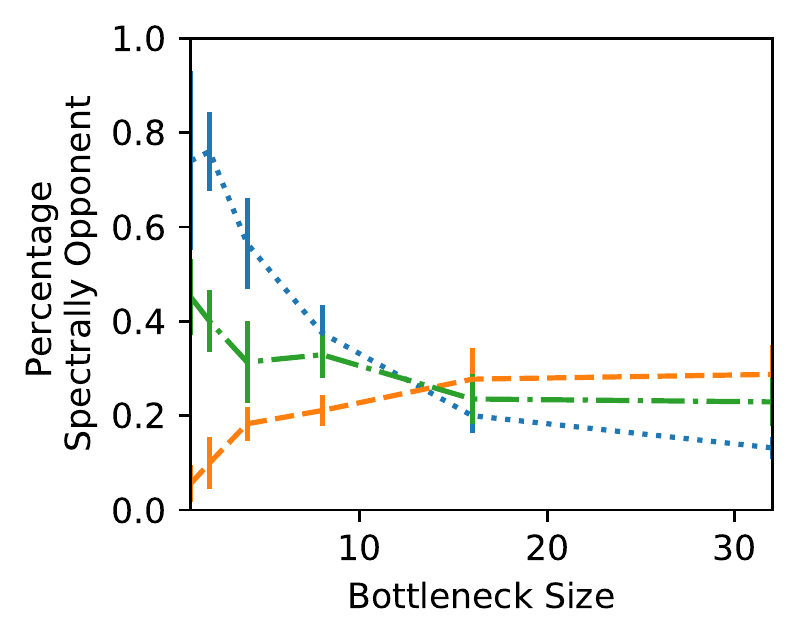}
        \caption{Spectrally Opponent}\label{fig:devalois:opponent}
    \end{subfigure}
    \begin{subfigure}{0.32\linewidth}
        \includegraphics[width=\linewidth]{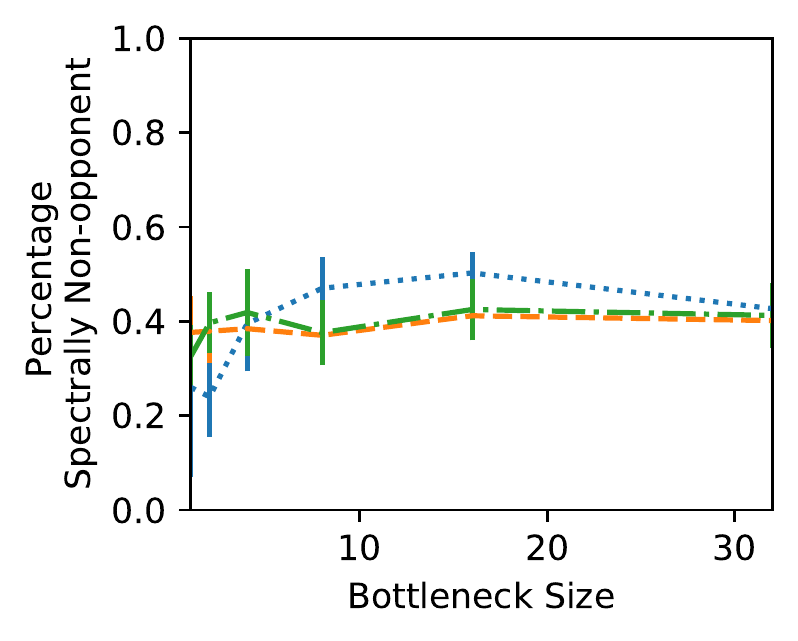}
        \caption{Non-opponent}\label{fig:devalois:nonopponent}
    \end{subfigure}
    \begin{subfigure}{0.32\linewidth}
        \includegraphics[width=\linewidth]{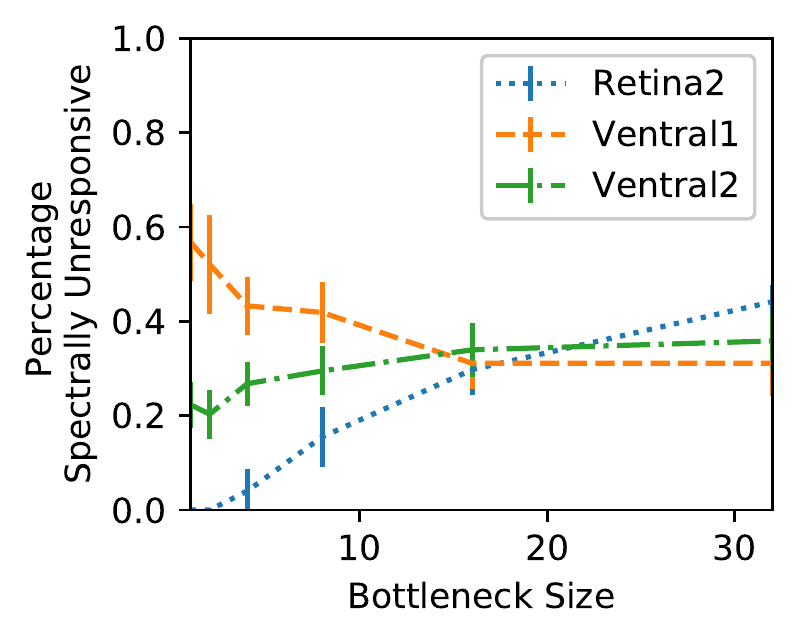}
        \caption{Unresponsive}\label{fig:devalois:unresponsive}
    \end{subfigure}
    \caption{Distribution of spectrally opponent, non-opponent and unresponsive cells in different layers of our model as a function of bottleneck size.}
    \label{fig:devalois}
\end{figure}
Curves showing how the distributions of the spectral classes change for the second retinal and first two ventral layers as the bottleneck is increased are given in Figure~\ref{fig:devalois}. As the bottleneck decreases, the second retina layer exhibits a strong increase in spectral opponency, nearing $100\%$ for a bottleneck of one. Conversely, cells in the first ventral layer show a decrease in spectral opponency over the same region.
Interestingly, for all but the tightest bottlenecks, up to half of the cells are spectrally non-opponent. Spectrally unresponsive cells show almost the exact opposite pattern to spectrally opponent cells.
The plots in Figure~\ref{fig:colours} from the Appendix show the distribution over the hue wheel of the most excitatory and most inhibitory colours in spectrally opponent cells for our models. Strikingly, the most common form of spectral opponency in the second retinal layer is decidedly red-green, corresponding well with the observable colour opponents in biological vision \citep{pridmore2011complementary}. In particular, the presence of cyan and magenta aligns well with observations from complementary colour theory \citep{pridmore2005theory}.
\paragraph{Spatial opponency}
%
To explore spatial opponency, we can use a similar set-up to our experiments with spectral opponency, measuring cell response to a series of high contrast greyscale gratings produced from a sinusoidal function for a range of rotations, frequencies and phases following \citet{johnson2001spatial} (see Appendix~\ref{app:gratings} for an example). We can subsequently classify a cell as spatially opponent, non-opponent or unresponsive by comparing the maximum and minimum responses against the baseline in the same way as before. Further, we can characterise whether a cell is orientation tuned by isolating the grating frequency which gives the largest response for all orientations and phases, then computing the average response per orientation for that frequency across all phases.
\begin{figure}
    \centering
    \begin{subfigure}{0.32\linewidth}
        \includegraphics[width=\linewidth]{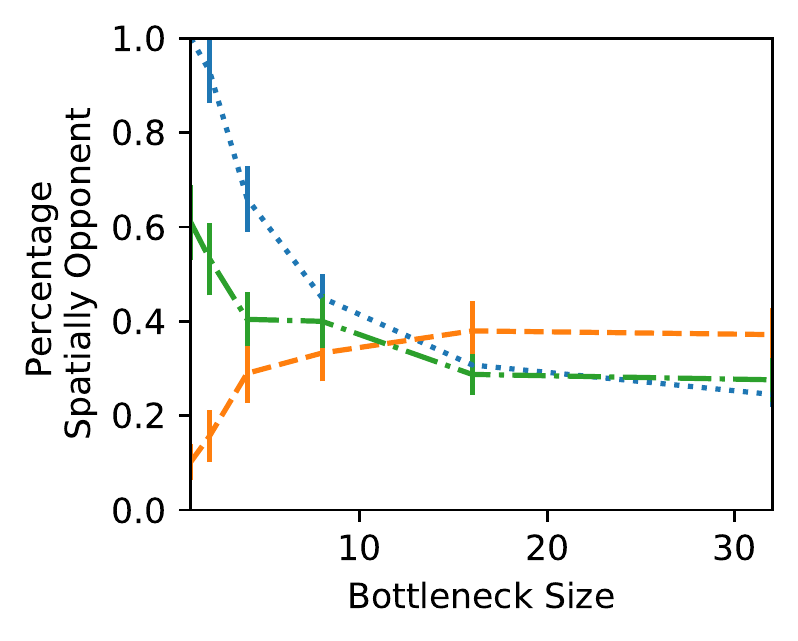}
        \caption{Spatially Opponent}\label{fig:spatial:opponent}
    \end{subfigure}
    \begin{subfigure}{0.32\linewidth}
        \includegraphics[width=\linewidth]{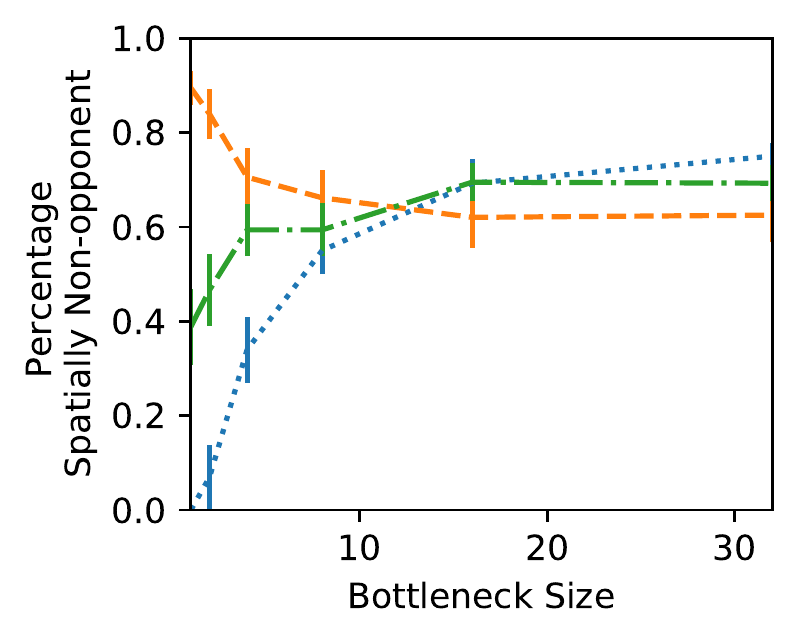}
        \caption{Non-opponent}\label{fig:spatial:nonopponent}
    \end{subfigure}
    \begin{subfigure}{0.32\linewidth}
        \includegraphics[width=\linewidth]{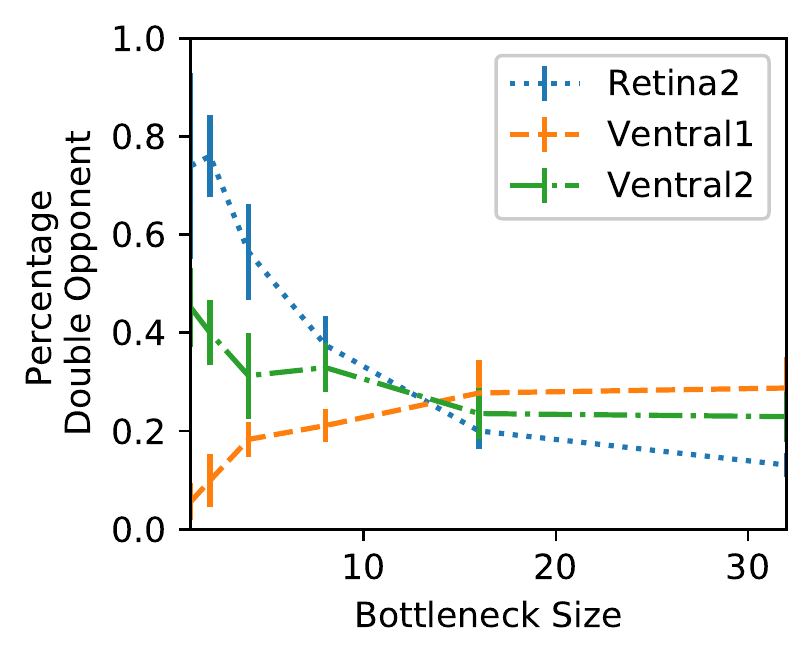}
        \caption{Double Opponent}\label{fig:double}
    \end{subfigure}
    \caption{Distribution of spatially opponent, non-opponent and double opponent cells in different layers of our model as a function of bottleneck size.}
    \label{fig:spatial}
\end{figure}
Automating the classification of cells, we can measure how spatial opponency manifests itself across the network, obtaining the results depicted in Figures~\ref{fig:spatial:opponent} and \ref{fig:spatial:nonopponent}. We omit spatially unresponsive cells as the percentages found in each layer was always on or very near zero. For a small bottleneck, the vast majority of cells in the second retinal layer are spatially opponent. Conversely, cells in the first ventral layer are predominantly spatially non-opponent. For less constrained bottlenecks these distributions converge to be approximately equal in each of the layers. Surprisingly, and contrasting with spectral opponency, almost all cells respond to some configuration of the grating stimulus, with only a small fraction of the population being spatially unresponsive. What is again clear in these experiments is the emergent structure that arises from the introduction of a bottleneck into the model.
\paragraph{Double opponency}
%
Following \citet{SHAPLEY2011701}, we can automatically classify a cell as being double opponent if it is both spectrally and spatially opponent. Figure \ref{fig:double} shows the distribution of double opponent cells as a function of bottleneck size, giving a similar picture to the spectral and spatial opponency plots. This finding is in alignment with biological observations that most spectrally opponent cells are orientation selective for both achromatic and chromatic stimuli \citep{johnson2008orientation}.
\paragraph{Generalisation to CIELAB space}
\begin{figure}
    \centering
    \begin{subfigure}{0.32\linewidth}
        \includegraphics[width=\linewidth]{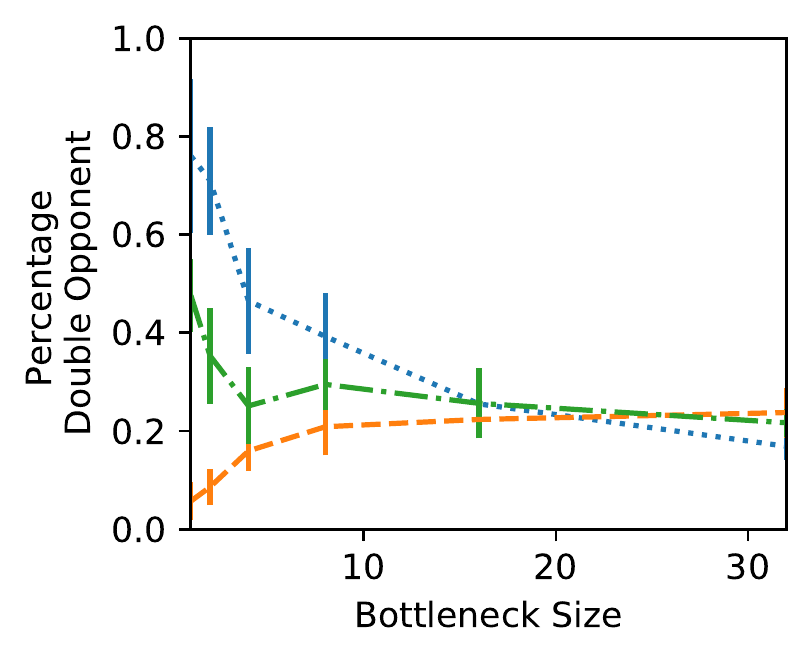}
        \caption{CIELAB: Double}\label{fig:double-lab}
    \end{subfigure}
    \begin{subfigure}{0.32\linewidth}
        \includegraphics[width=\linewidth]{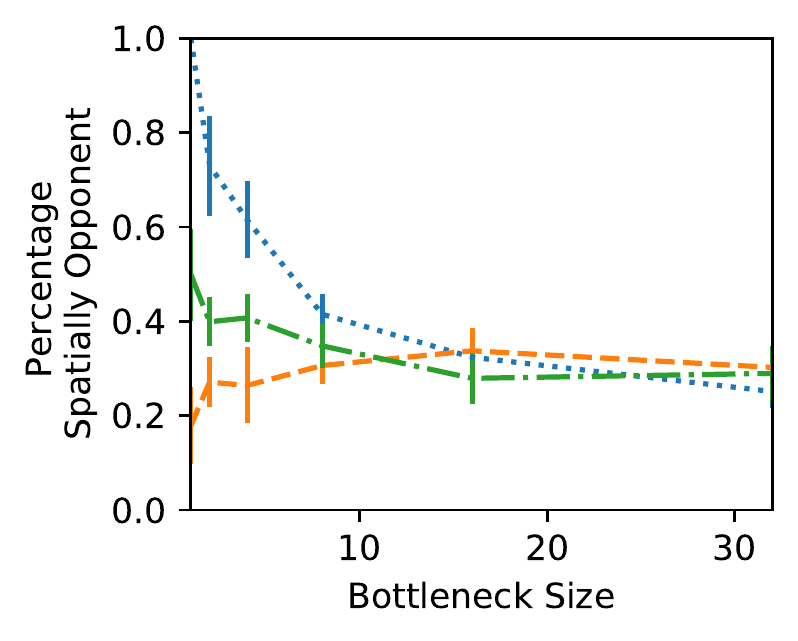}
        \caption{Shuffled: Spatial}\label{fig:spatial:shuffled}
    \end{subfigure}
    \begin{subfigure}{0.32\linewidth}
        \includegraphics[width=\linewidth]{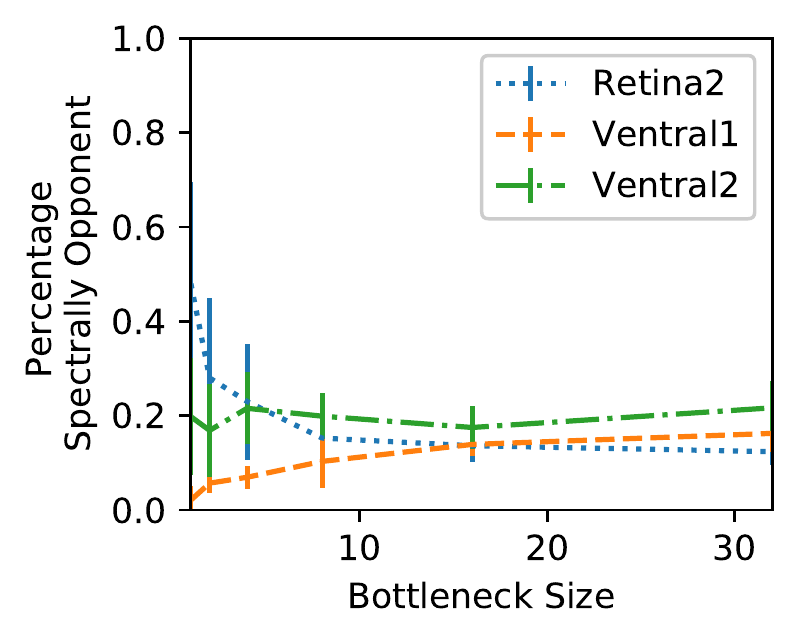}
        \caption{Shuffled: Spectral}\label{fig:devalois:shuffled}
    \end{subfigure}
    \caption{ (\protect\subref{fig:double-lab}) Distribution of double opponent cells in different layers of our model trained on CIELAB images as a function of bottleneck size and the effect of shuffling the colour channels has on spatial opponency (\protect\subref{fig:spatial:shuffled}) and spectral opponency (\protect\subref{fig:devalois:shuffled}).}
\end{figure}
We performed additional experiments to validate whether double opponency is still a feature in networks trained on images in CIELAB space. 
Figure \ref{fig:double-lab} shows the distribution of double opponent cells in this setting. As a strong validation of our findings, the distribution is nearly identical to that of networks trained on images in RGB space. This again supports the suggestion that double opponent characteristics arise from the statistics of natural images.
\paragraph{Ablation: ventral depth}
In order to build a greater understanding of the conditions required for opponency, we plot the distribution of spectrally and spatially opponent cells as a function of ventral depth in Figures~\ref{fig:devalois-depth} and \ref{fig:spatial-depth} from the Appendix.
With each added layer, the same pattern is found, shifted one layer to the right. This suggests that the functional organisation depends more on the distance from the output layer than from the input.
\paragraph{Ablation: spectral consistency}
For a final controlled demonstration of the conditions required for double opponency, we remove colour information by randomly shuffling the colour channels of inputs to the network. The resultant distribution plots show that this alteration completely removes spectral opponency (Figure~\ref{fig:devalois:shuffled}), whilst spatial opponency remains (Figure~\ref{fig:spatial:shuffled}). This again strengthens the observation that opponent characteristics arise as a result of the statistics of natural images.
\section{Discussion}
Our investigation has shown that a dimensionality bottleneck has the power to do more than just induce centre-surround receptive fields in the retina and oriented receptive fields in V1. We have shown that spectral, spatial and double opponent characteristics arise from this constriction and made two key observations.
First, we have shown that a retinal bottleneck induces structure in the network where all cells in each layer follow a layer dependant functional archetype. Second, we have given a strong demonstration that opponent cells emerge as a result of the statistics of the input space; in this case, natural images. Our findings also have the potential to support the development of new network architectures. Specifically, if one accepts that human superiority in visual problems (such as adversarial robustness or constructing a notion of shape) can be approached through increasing similarity between deep networks and the human visual system, then we have provided a strong mandate for future research into the use of convolutional bottlenecks.

\bibliographystyle{plainnat}
\bibliography{main}

\appendix
\appendixpage

\section{Model Details}\label{app:details}

In this appendix we provide further exposition of the details of our model and experimentation process. In \ref{app:colorpatches} we detail our process for generating colour patches and obtaining a spectral tuning curve for a particular neuron. In \ref{app:gratings} we give example gratings images used in our spatial experiments.

\subsection{Model Training}\label{app:training}
\begin{figure}[h!]
    \centering
    \begin{subfigure}{0.49\linewidth}
        \includegraphics[width=\linewidth]{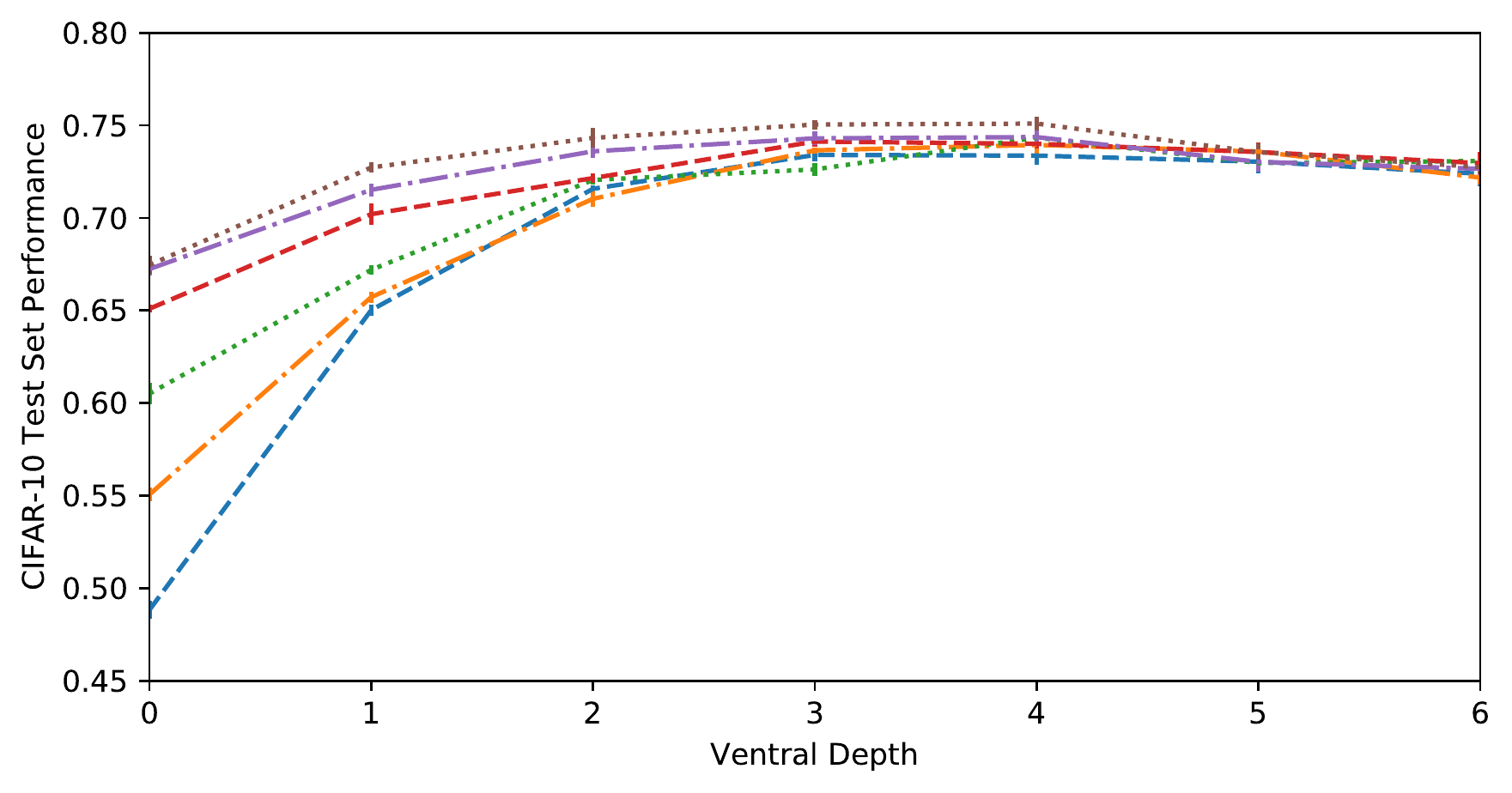}
        \caption{Greyscale}
    \end{subfigure}
    \begin{subfigure}{0.49\linewidth}
        \includegraphics[width=\linewidth]{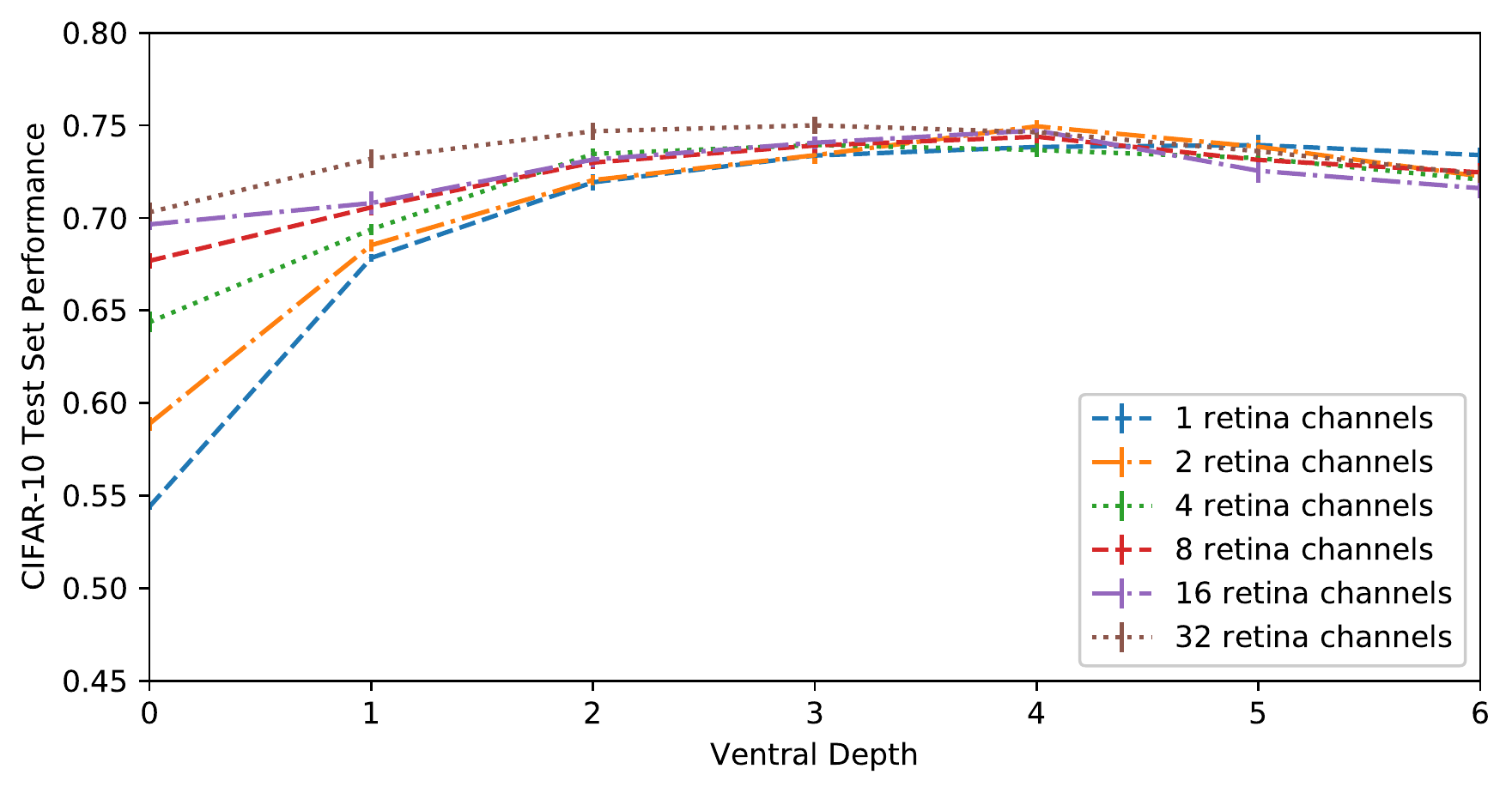}
        \caption{Colour}
    \end{subfigure}
    \caption{Test accuracy for the different combinations of retinal bottleneck  and ventral stream depth explored in the experiments. Data points are the average over 10 trials.}
    \label{fig:model-acc}
\end{figure}

We exactly follow the model construction and training procedure defined by \citet{lindsey2019unified}. We note in addition that to replicate the results of the original experiments (see below). We additionally use a weight decay of $1e-6$ to provide mild regularisation of the networks weights, and data augmentation (random translations of 10\% of the image width/height, and random horizontal flipping) to avoid over-fitting. Figure~\ref{fig:model-acc} gives the average terminal accuracy for models trained both on greyscale and colour images. The greyscale accuracy curves match those given in \citet{lindsey2019unified}.

\subsection{Gratings}\label{app:gratings}
\begin{figure}[h!]
    \centering
    \begin{subfigure}{0.15\linewidth}
        \includegraphics[width=\linewidth]{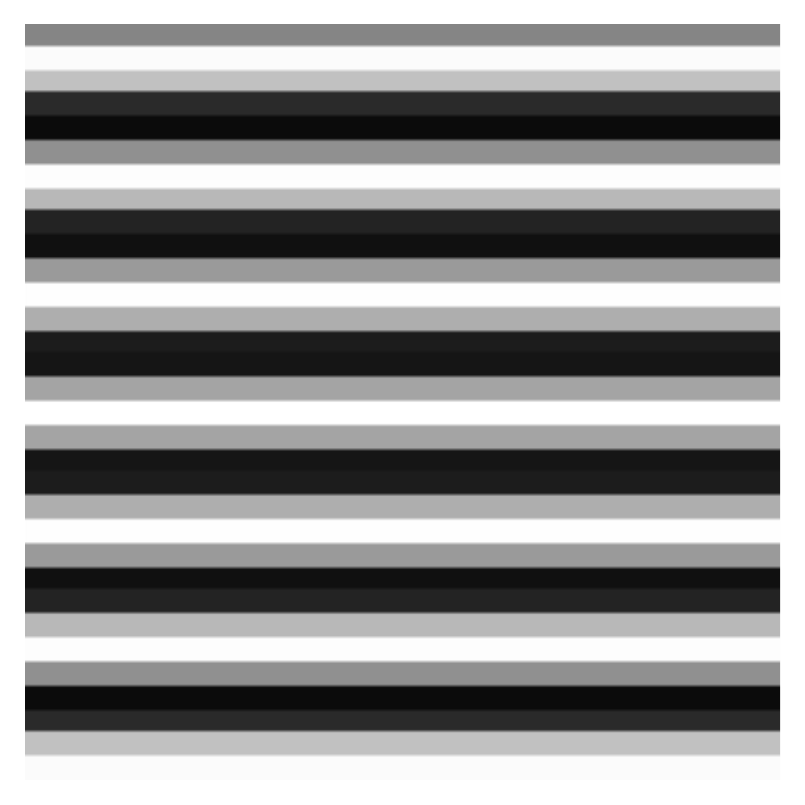}
        \caption{$\theta=\ang{0}$}
    \end{subfigure}
    \hspace{0.05\linewidth}
    \begin{subfigure}{0.15\linewidth}
        \includegraphics[width=\linewidth]{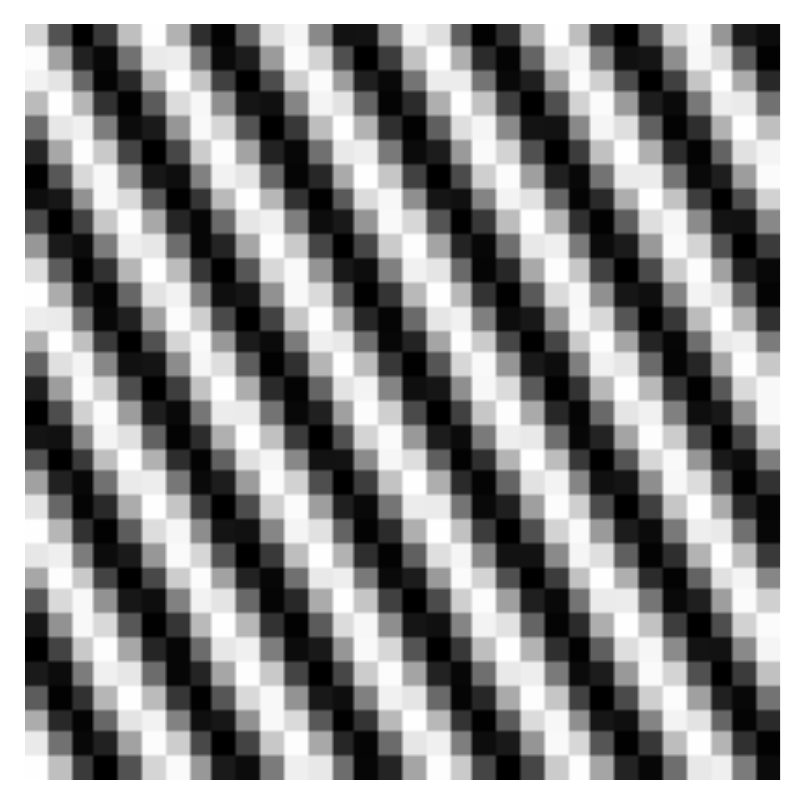}
        \caption{$\theta=\ang{45}$}
    \end{subfigure}
    \hspace{0.05\linewidth}
    \begin{subfigure}{0.15\linewidth}
        \includegraphics[width=\linewidth]{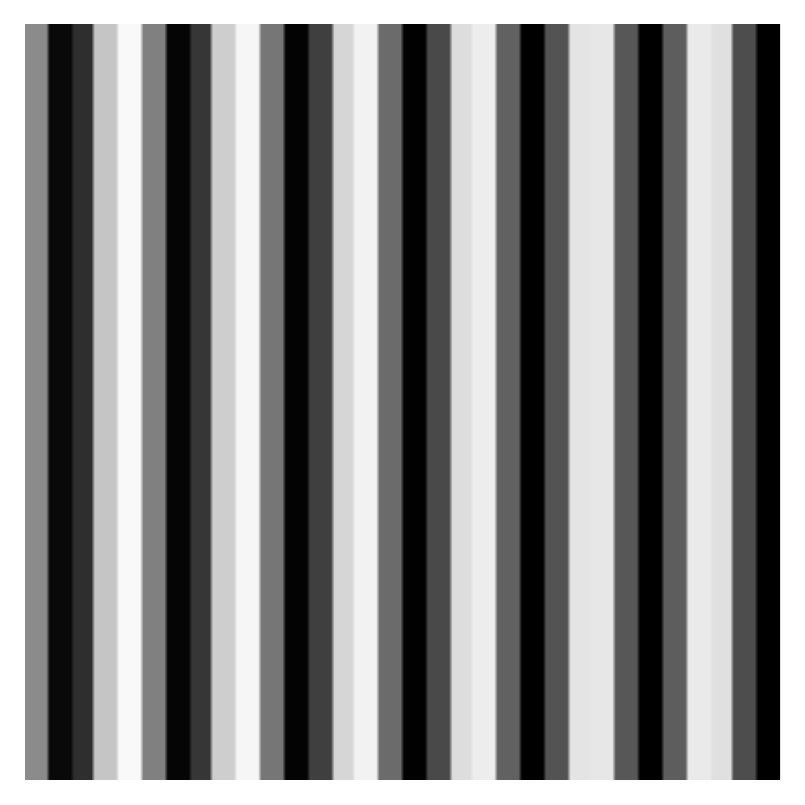}
        \caption{$\theta=\ang{90}$}
    \end{subfigure}
    \hspace{0.05\linewidth}
    \begin{subfigure}{0.15\linewidth}
        \includegraphics[width=\linewidth]{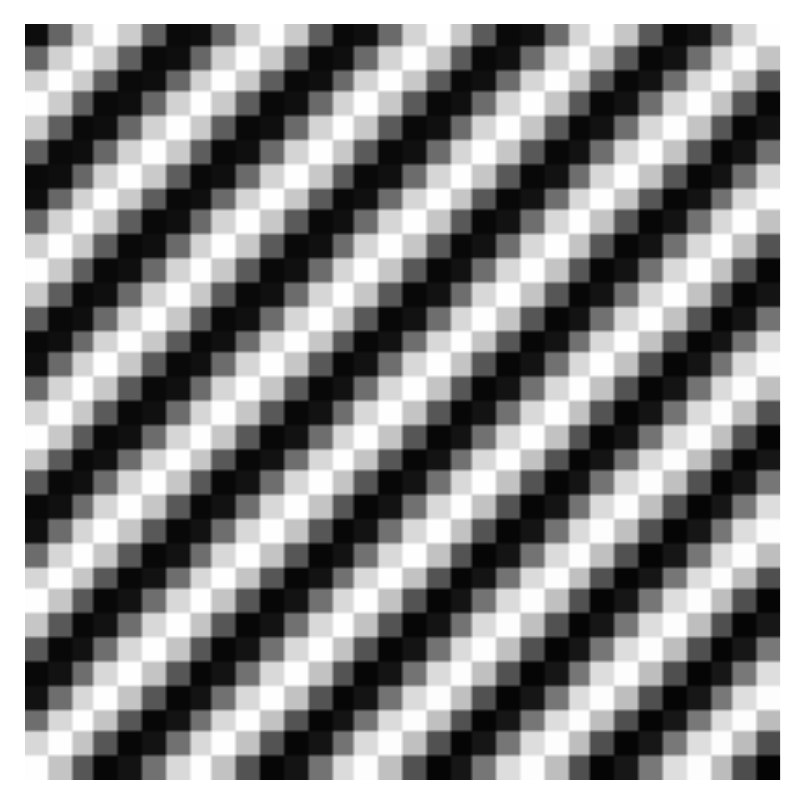}
        \caption{$\theta=\ang{135}$}
    \end{subfigure}
    \caption{Examples of grating patterns used as stimuli for the spatial opponency experiments.}
    \label{fig:greygratings}
\end{figure}

The grating patterns in Figure \ref{fig:greygratings} illustrate the type of stimuli used to classify cells according to their orientation and form sensitivity. These samples have been generated with different angles ($\theta$), frequency of $\frac{8}{2\pi}$ and $\ang{0}$ phase. For the experiments described in Section 4.2 of the paper, a wide range of values has been used to generate this type of stimuli.

\subsection{Sampling Colour Patches}\label{app:colorpatches}
\begin{figure}[h!]
    \centering
    \begin{subfigure}{\linewidth}
        \includegraphics[width=\linewidth]{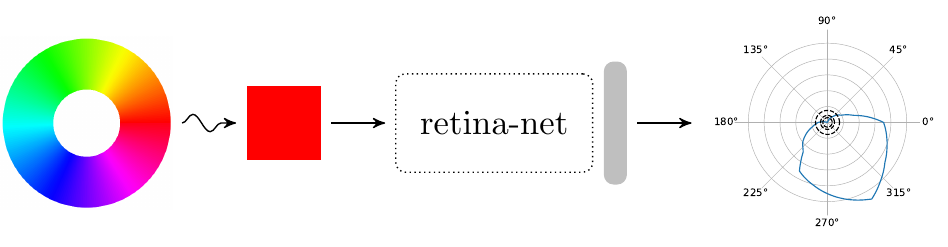}
        \caption{Spectral opponency}
    \end{subfigure}
    \begin{subfigure}{\linewidth}
        \includegraphics[width=\linewidth]{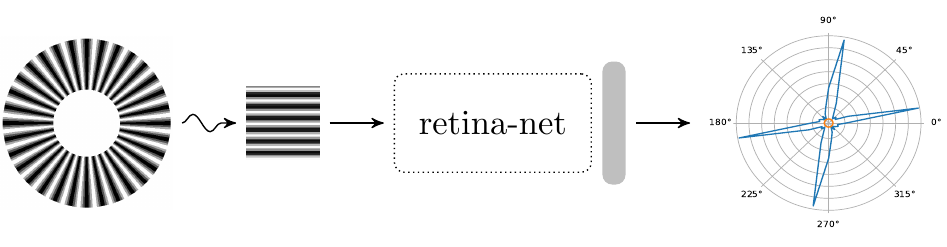}
        \caption{Spatial opponency}
    \end{subfigure}
    \caption{Our approach for characterising spectral and spatial opponency. Uniform colour patches of differing hue or gratings of differing frequency, phase and angle are sampled and presented to the network. Responses are measured for a cell across all inputs to construct a response curve (blue line in the right-hand plots).}
    \label{fig:model}
\end{figure}

Our process for extracting tuning curves is shown in Figure~\ref{fig:model}. For our spectral experiments, the stimuli provided to the network were colour patches sampled from the HSL colour space by varying the hue. All input colours have full saturation (i.e. $S=1$) and are equiluminant (i.e. $L=0.5$).
We used the following conversion formula to transition between HSL and RGB colour spaces:

\begin{align}
    (R,G,B) &= ((R^\prime +m)\times 255, (G^\prime +m)\times 255,(B^\prime+m)\times 255)
\end{align}
where $(R^\prime, G^\prime, B^\prime)$ is defined as
\begin{align}
   (R^\prime, G^\prime, B^\prime) = 
    \begin{cases}
        (C,X,0) & \text{for }\ang{0}\le H<\ang{60}\\    
        (X,C,0) & \text{for }\ang{60}\le H<\ang{120}\\
        (0,C,X) & \text{for }\ang{120}\le H<\ang{180}\\
        (0,X,C) & \text{for }\ang{180}\le H<\ang{240}\\
        (X,0,C) & \text{for }\ang{240}\le H<\ang{300}\\
        (C,0,X) & \text{for }\ang{300}\le H<\ang{360}
    \end{cases}
\end{align}

$C,X \text{and } m$ are calculated as follows with the set $S$ and $L$ values and $H$ ranging from $\ang{0}$ to $\ang{360}$:
\begin{align}
\begin{split}
    C &=(1-|2L-1|) \times S\\
    X &=C \times (1-|(H/\ang{60}) \textrm{ mod } 2 - 1|)\\
    m &= L - C/2
\end{split}
\end{align}

\section{Ventral Depth}
\begin{figure}[h!]
    \centering
    \begin{subfigure}{0.32\linewidth}
        \includegraphics[width=\textwidth]{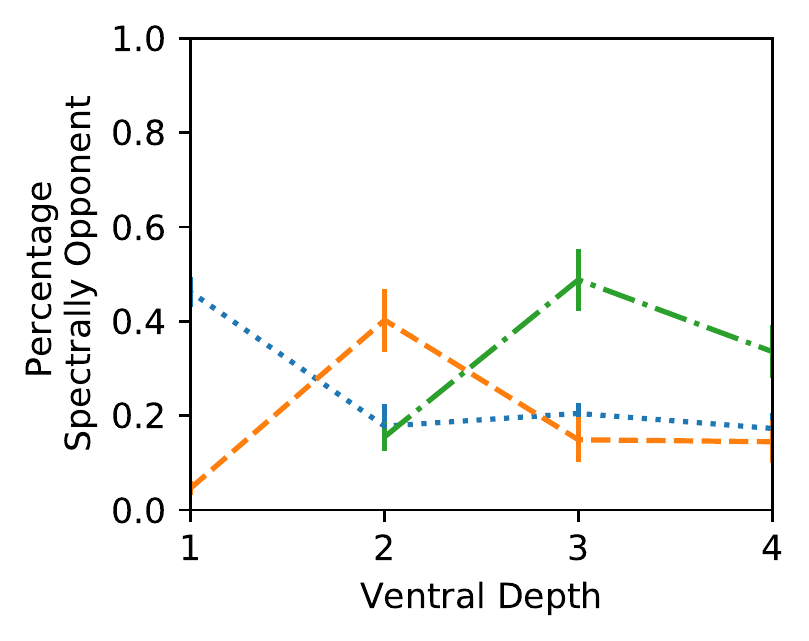}
        \caption{Spectrally Opponent}
    \end{subfigure}
    \begin{subfigure}{0.32\linewidth}
        \includegraphics[width=\textwidth]{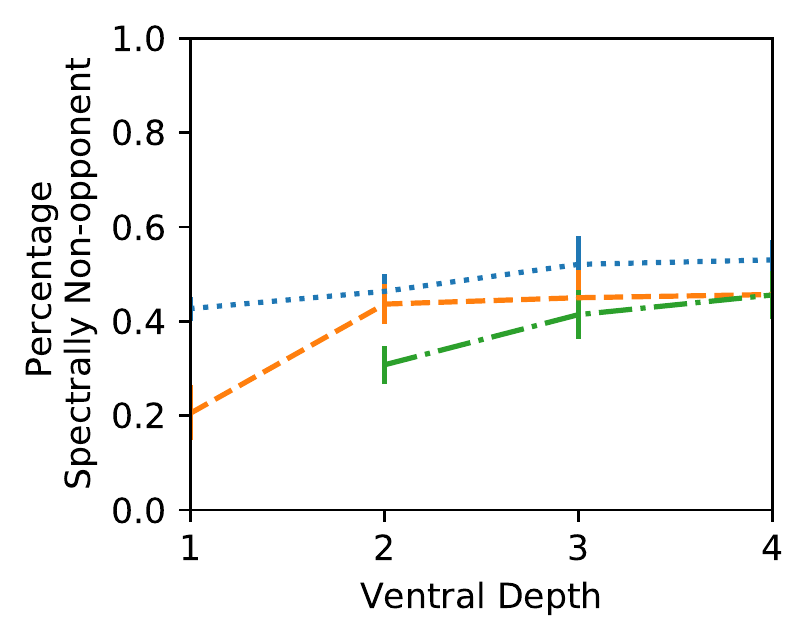}
        \caption{Non-opponent}
    \end{subfigure}
    \begin{subfigure}{0.32\linewidth}
        \includegraphics[width=\textwidth]{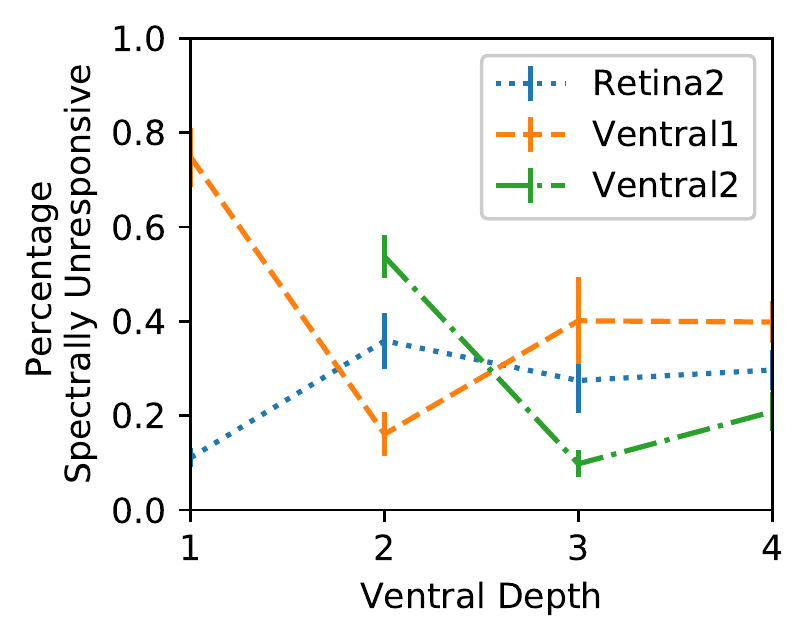}
        \caption{Unresponsive}
    \end{subfigure}
    \caption{Distribution of spectrally opponent, non-opponent and unresponsive cells in different layers of our model following the definitions given by \citet{de1958response} as a function of ventral depth.}
    \label{fig:devalois-depth}
\end{figure}

\begin{figure}[h!]
    \centering
    \begin{subfigure}{0.32\linewidth}
        \includegraphics[width=\textwidth]{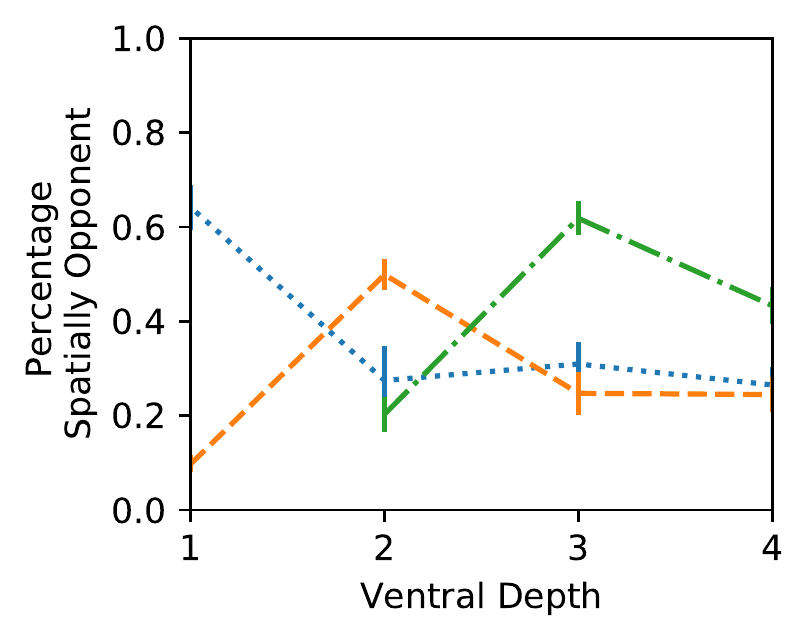}
        \caption{Spatially Opponent}
    \end{subfigure}
    \begin{subfigure}{0.32\linewidth}
        \includegraphics[width=\textwidth]{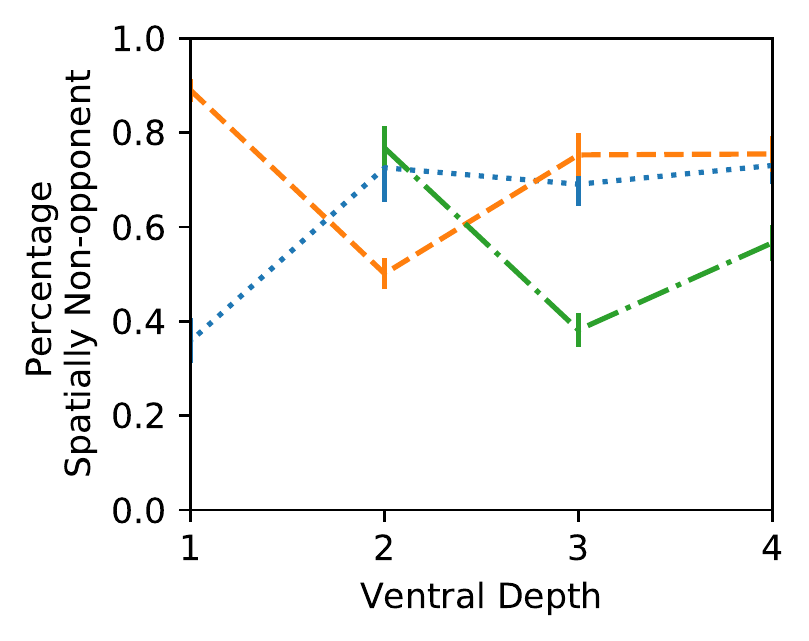}
        \caption{Non-opponent}
    \end{subfigure}
    \begin{subfigure}{0.32\linewidth}
        \includegraphics[width=\textwidth]{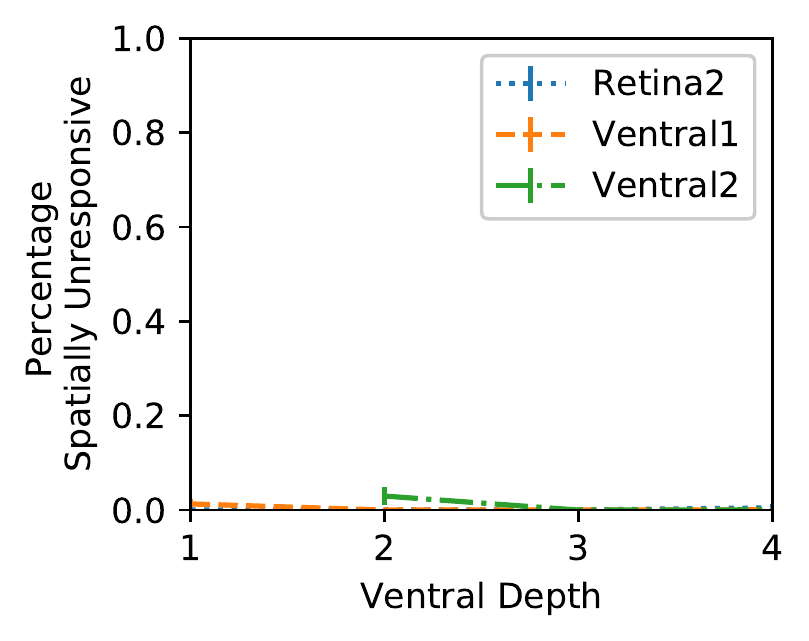}
        \caption{Unresponsive}
    \end{subfigure}
    \caption{Distribution of spatially opponent, non-opponent and unresponsive cells in different layers of our model following the definitions given by \citet{johnson2001spatial} as a function of ventral depth.}
    \label{fig:spatial-depth}
\end{figure}

\clearpage

\section{Distribution of Colours}

\begin{figure}[h!]
    \centering
    \begin{subfigure}{0.25\linewidth}
        \includegraphics[width=\textwidth]{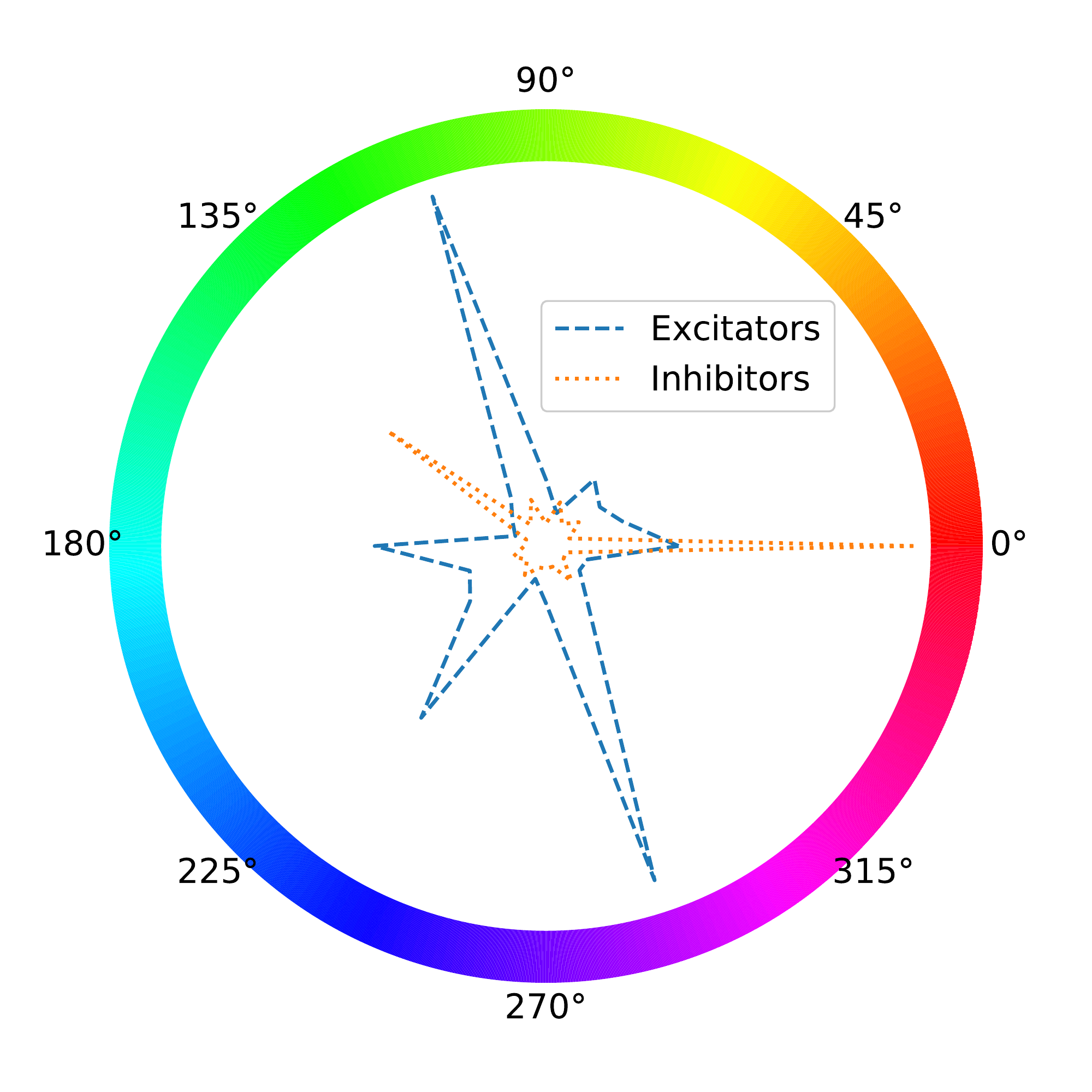}
        \caption{Retina 2}
    \end{subfigure}
    \hspace{0.1\linewidth}
    \begin{subfigure}{0.25\linewidth}
        \includegraphics[width=\textwidth]{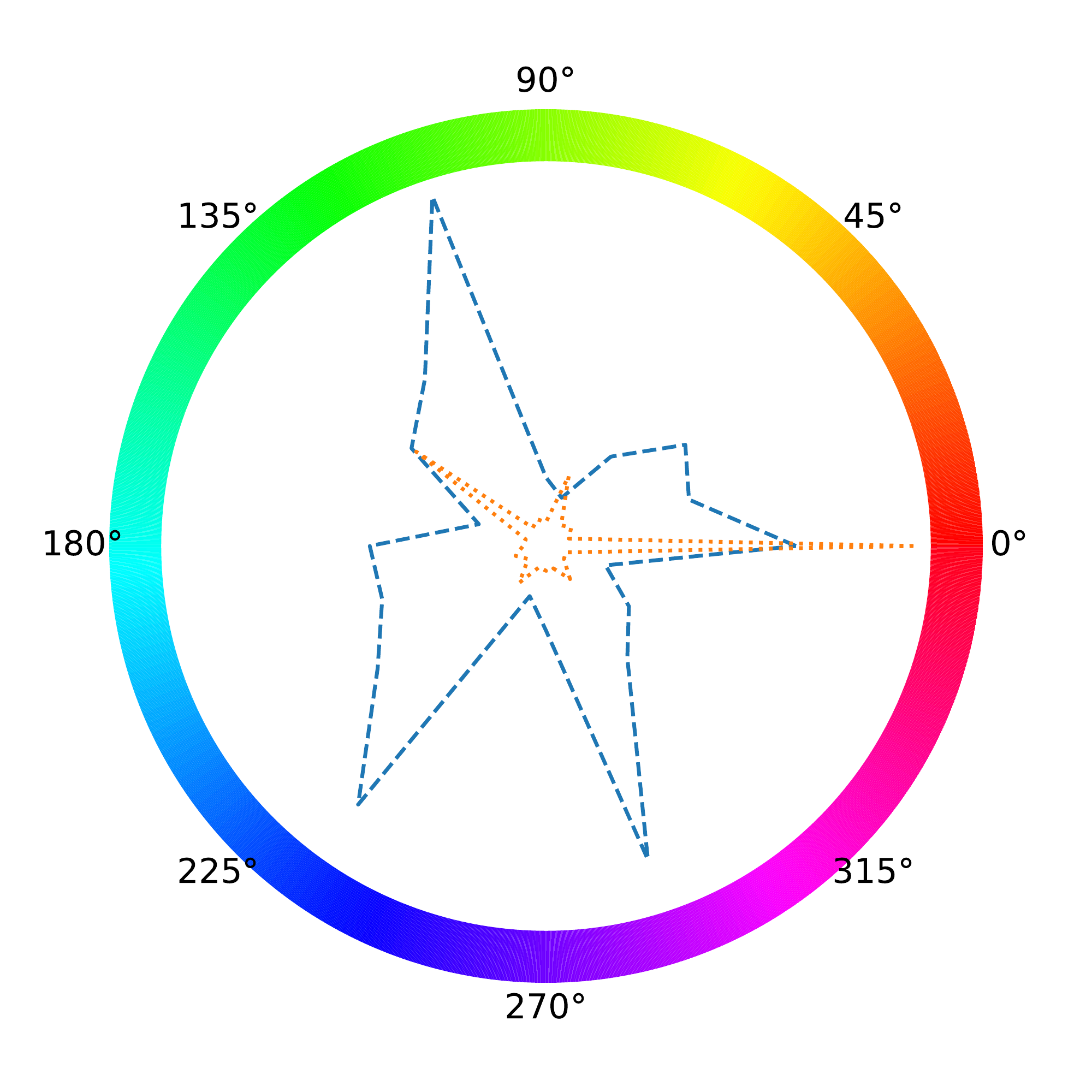}
        \caption{Ventral 1}
    \end{subfigure}
    \hspace{0.1\linewidth}
    \begin{subfigure}{0.25\linewidth}
        \includegraphics[width=\textwidth]{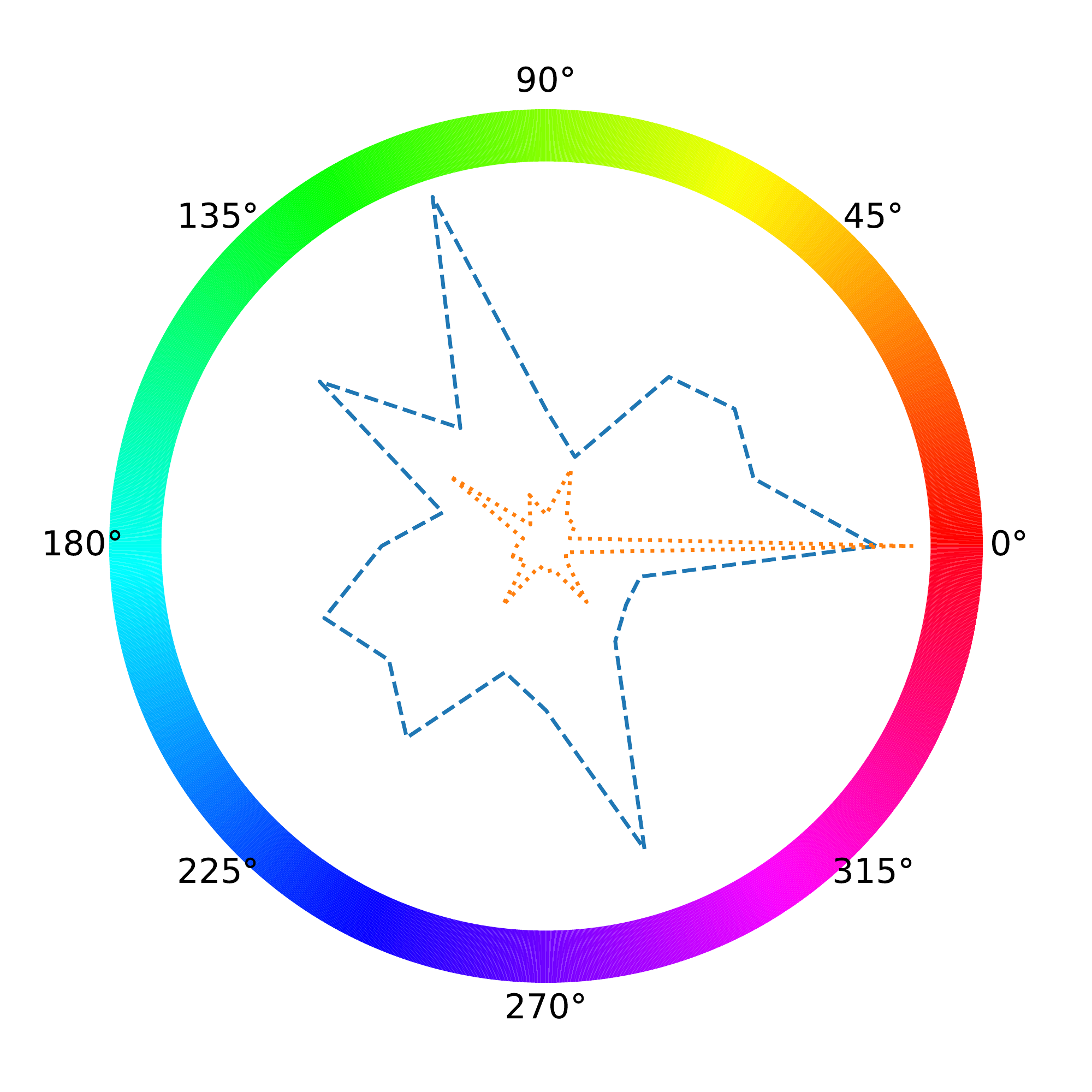}
        \caption{Ventral 2}
    \end{subfigure}
    \caption{Distribution of most excitatory and most inhibitory colours for spectrally opponent cells at different depths in our model.}
    \label{fig:colours}
\end{figure}


\section{Characterising a Single Cell}\label{app:single}
\begin{figure}[h!]
    \centering
    \begin{subfigure}{\linewidth}
        \includegraphics[width=\linewidth]{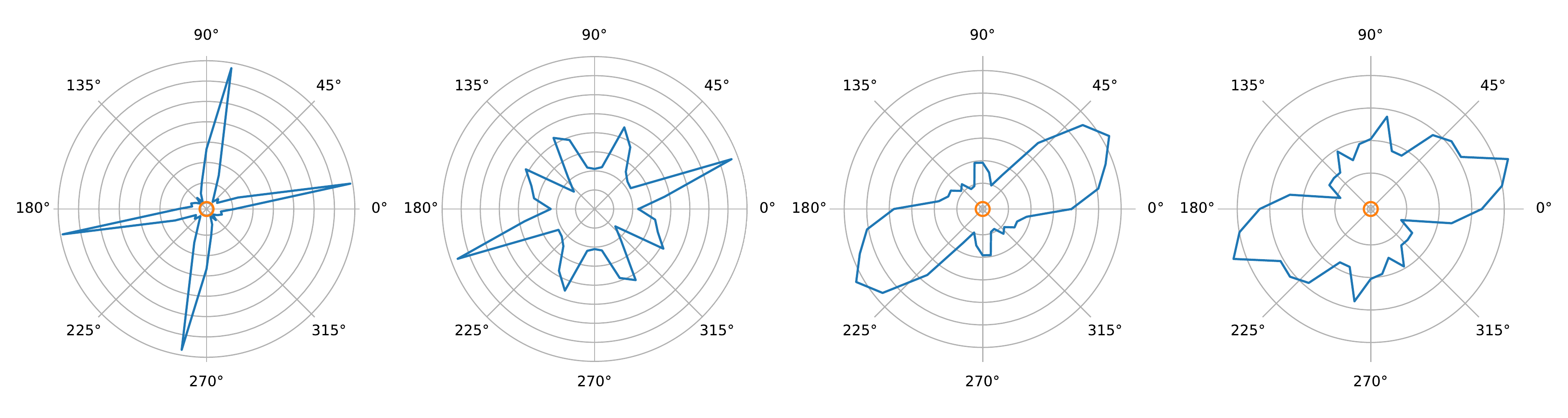}
        \caption{Orientation Tuning Curve}
        \label{subfig:spatial}
    \end{subfigure}
    \begin{subfigure}{\linewidth}
        \includegraphics[width=\linewidth]{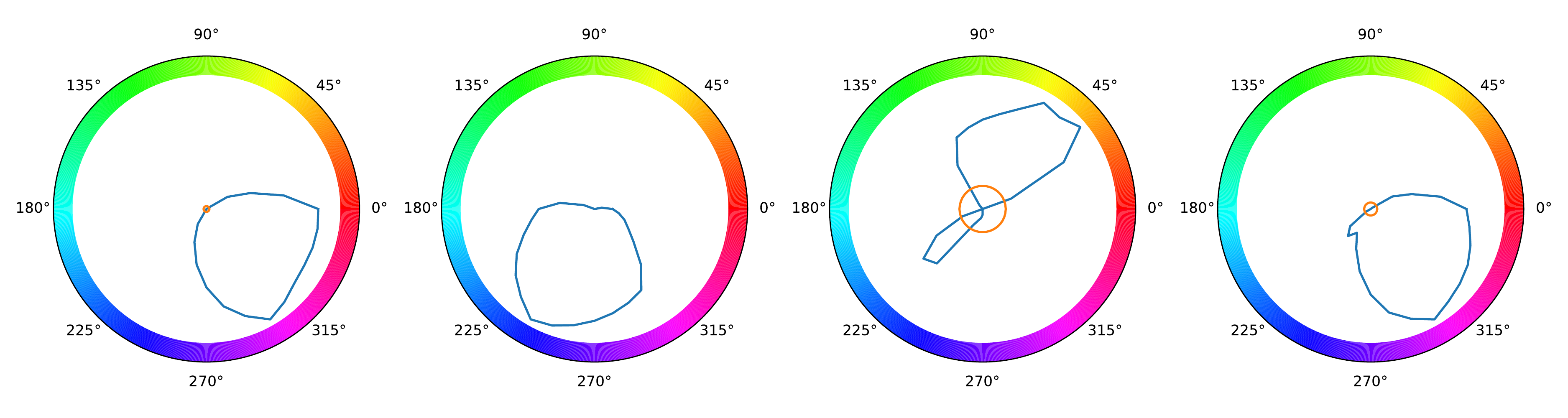}
        \caption{Spectral Tuning Curve}
        \label{subfig:spect}
    \end{subfigure}
    \begin{subfigure}{\linewidth}
        \includegraphics[width=\linewidth]{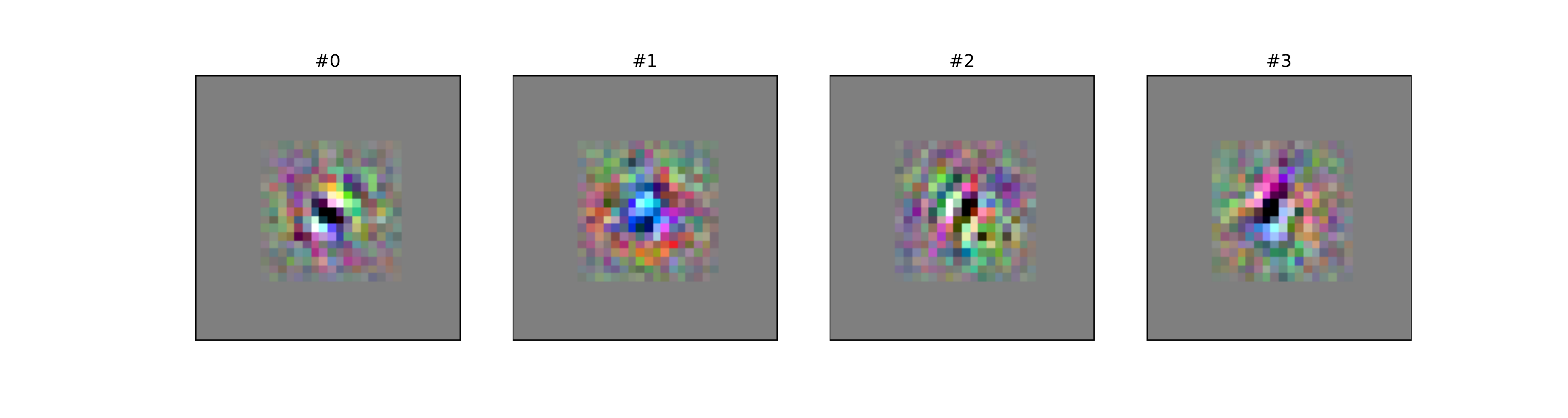}
        \caption{First Order RF Approximation}
    \end{subfigure}
    \caption{Characterisation of the 4 cells in the second retinal layer of a network with bottleneck of 4 and ventral depth of 2, based on (a) orientation and form sensitivity to a range of grating patterns, (b) colour sensitivity to the colour stimuli shown on the Hue wheel, and (c), the receptive field approximation via 1-step gradient ascent towards a blank image.}
    \label{fig:cellcharac}
\end{figure}

\end{document}